\documentclass[10pt,twocolumn,letterpaper]{article}
\usepackage{cvpr} 

\usepackage[ruled,vlined,linesnumbered]{algorithm2e}
\DontPrintSemicolon                
\SetAlgoVlined                     
\SetInd{0.25em}{0.8em}             
\SetAlgoInsideSkip{0.2em}          
\SetAlgoSkip{0.4em}                

\usepackage{mathtools}

\DeclareMathOperator*{\argmax}{arg\,max}
\DeclareMathOperator*{\argmin}{arg\,min}

\usepackage{enumitem}

\definecolor{cvprblue}{rgb}{0.21,0.49,0.74}
\usepackage[pagebackref,breaklinks,colorlinks,allcolors=cvprblue]{hyperref}
\usepackage{makecell}
\usepackage[table]{xcolor}
\definecolor{rowgray}{gray}{0.92}
\usepackage{pgfplots}
\pgfplotsset{compat=1.18}
\usepackage{booktabs}
\usepackage{multirow}
\usepackage{graphicx}   
\newcommand{\modelgroup}[2]{%
  \multirow{#1}{*}{%
    \rotatebox[origin=c]{90}{\scriptsize\bfseries #2}%
  }%
}

\usepackage{array}      
\def\paperID{3259} 
\def\confName{CVPR}
\def\confYear{2026}

\title{ReDiPrune: Relevance-Diversity Pre-Projection Token Pruning for Efficient Multimodal LLMs}

\author{
An Yu\\
University at Albany, SUNY\\
\and
Ting Yu Tsai\\
University at Albany, SUNY\\
\and Zhenfei Zhang\\
University at Albany, SUNY\\
\and
Weiheng Lu\\
Peking University\\
\and 
Felix X.-F. Ye\\
University at Albany, SUNY\\
\and 
Ming-Ching Chang\\
University at Albany, SUNY\\
}

\begin{document}
\maketitle

  
 Recent multimodal large language models are computationally expensive because Transformers must process a large number of visual tokens. We present \textbf{ReDiPrune}, a training-free token pruning method applied \emph{before} the vision-language projector, where visual features remain rich and discriminative. Unlike post-projection pruning methods that operate on compressed representations, ReDiPrune selects informative tokens directly from vision encoder outputs, preserving fine-grained spatial and semantic cues. Each token is scored by a lightweight rule that jointly consider text-conditioned relevance and max-min diversity, ensuring the selected tokens are both query-relevant and non-redundant. ReDiPrune is fully plug-and-play, requiring no retraining or architectural modifications, and can be seamlessly inserted between the encoder and projector. Across four video and five image benchmarks, it consistently improves the accuracy-efficiency trade-off. For example, on EgoSchema with LLaVA-NeXT-Video-7B, retaining only 15\% of visual tokens yields a +2.0\% absolute accuracy gain while reducing computation by more than $6\times$ in TFLOPs. Code is available at \url{https://github.com/UA-CVML/ReDiPrune}.    

\begin{figure}[t]
\centerline{
    \includegraphics[width=0.5\textwidth]{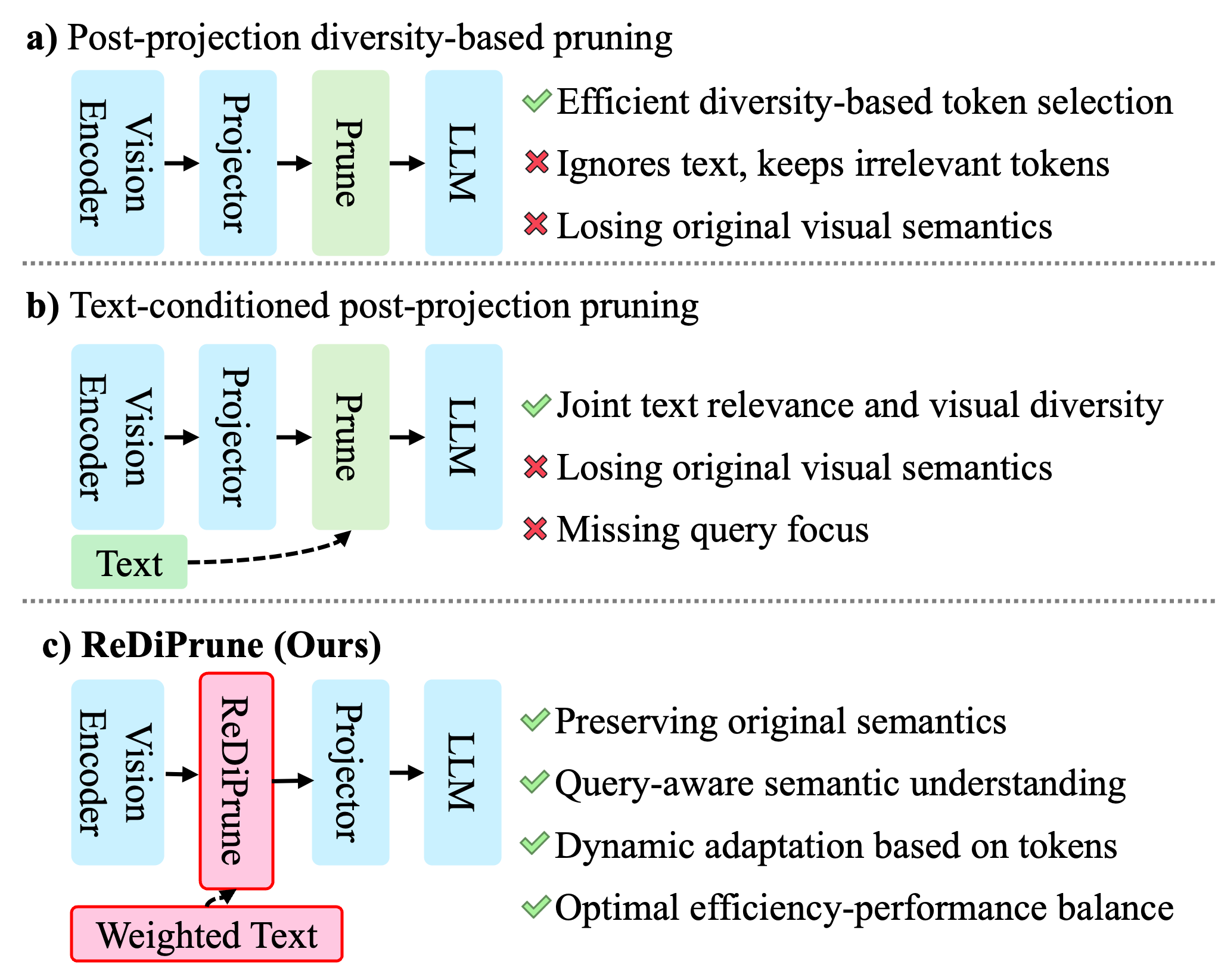}
    \vspace{-2mm}
}  
\caption{
\textbf{Comparison of pruning strategies in MLLMs:}
(a) Post-projection pruning selects diverse tokens but ignores text and loses fine details.
(b) Text-guided post-projection pruning improves query relevance but discards original visual cues.
(c) ReDiPrune prunes tokens {\em before projection} using weighted query embeddings, preserving details while balancing accuracy and efficiency.}
\vspace{-4mm}
\label{fig:teaser_figure}
\end{figure}


\section{Introduction}
\label{sec:intro}

Multimodal large language models (MLLMs)~\cite{videollava,llavavideo,videollama,wang2024videoagent} extend text-only LLMs to jointly process images, videos, and text, and have advanced rapidly in recent years. However, their efficiency remains limited by the quadratic cost of Transformer attention, with visual tokens contributing most of the computational overhead. Dense spatial patches and redundant visual content result in excessively long token sequences, leading to high memory consumption and slow inference.

In standard MLLM pipelines, visual inputs are encoded into dense tokens, projected into the LLM’s embedding space, and interleaved with text tokens. This process often exposes the LLM to thousands of visual tokens, significantly inflating sequence length, latency, and memory consumption, particularly for high-resolution images or long videos. To address this issue, token pruning~\cite{Alvar2025divprune,chen2024FastV,zhang2025CDPruner} has emerged as an effective strategy to reduce computation by removing redundant or less informative visual tokens.

Prior work~\cite{chen2024FastV,yang2024llavaprumerge,huang2025PruneVid} shows that pruning 50-95\% of visual tokens can preserve accuracy and improve efficiency, although methods that require calibration or fine-tuning are costly to adapt across architectures. Existing approaches differ primarily in \textbf{where} and \textbf{how} tokens are removed (Fig.~\ref{fig:teaser_figure}). Post-projection methods (Fig.~\ref{fig:teaser_figure}a) operate on the projected visual representations $Z_V=P(E_V)$.  Representative approaches include FastV~\cite{chen2024FastV}, PruMerge~\cite{yang2024llavaprumerge}, and VTW~\cite{lin2025VTW}, which rank or merge tokens to reduce redundancy, but often at the cost of fine-grained spatial details. DivPrune~\cite{Alvar2025divprune} maximizes diversity but ignores text, while CDPruner~\cite{zhang2025CDPruner} adds query conditioning yet still prunes after projection, where modality mixing can weaken original visual semantics. Pre-projection approaches such as TRIM~\cite{song2024TRIM} (Fig.~\ref{fig:teaser_figure}b) reduce tokens by ranking CLIP patches and retraining a LLaVA-1.5-TRIM model. While effective at moderate compression ($\approx$ 21\% tokens), TRIM depends on large-scale instruction tuning: each backbone must be retrained on 665K examples and performance degrades sharply under aggressive pruning (a 5\% keep ratio causes a 6-8 point drop across 12 benchmarks). Its reliance on CLIP and training further limits generality.

\begin{figure*}[t]
\centerline{
    \includegraphics[width=1.0\linewidth]{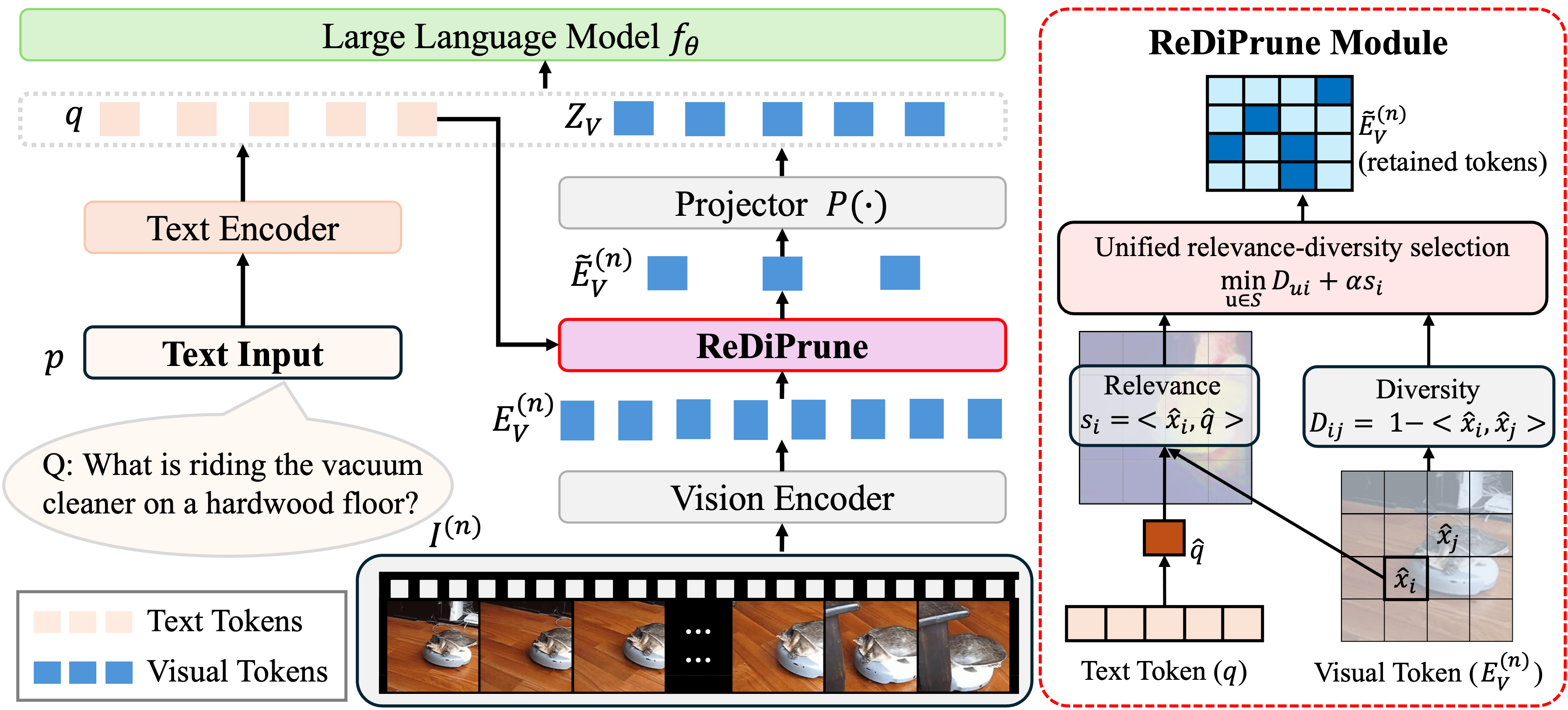}  
}
\caption{\textbf{Overview of ReDiPrune:} Given a prompt $p$, we build a normalized weighted query vector $\hat{q}$ ($\S$~\ref{sec:query_embedding}). For each frame $I^{(n)}$, the vision encoder produces tokens $E_V^{(n)}$. ReDiPrune operates \emph{before} the projector $P(\cdot)$ in the visual feature space, computing text relevance $s_i$ and cosine dissimilarity $D_{ij}$ ($\S$~\ref{sec:scoring}). With a per-frame budget $k_n$, we greedily select $k_n$ tokens by optimizing a relevance-diversity score $\min_{u\in S} D_{ui} + \alpha s_i$ ($\S$~\ref{sec:greedy_solver}), where $u$ indexes tokens already selected in $S$. The retained tokens $\tilde{E}_V^{(n)}$ are projected as $Z_V^{(n)}$ and concatenated with text embeddings for decoding by the LLM $f_\theta$, reducing redundancy while preserving query-relevant information.
}
\label{fig:architecture} 
\vspace{-2mm}
\end{figure*}

To address these limitations, we introduce ReDiPrune (Fig.~\ref{fig:teaser_figure}c, Fig.~\ref{fig:architecture}), a query-guided, training-free token selection module operating directly in the vision encoder feature space $E_V$. ReDiPrune scores visual tokens with a lightweight objective that combines text relevance and max-min diversity, selecting a semantically aligned and non-redundant subset. Pruning is performed prior to the vision–language projector, such that only the retained visual tokens are projected and forwarded to the LLM, thereby reducing end-to-end computation. Given an input prompt, we construct a weighted textual query, compute relevance scores for visual tokens in each frame, and greedily select tokens under a fixed budget. The selected tokens are then projected and interleaved with text embeddings for decoding. This upstream design avoids making selection decisions after cross-modal mixing and leads to tangible efficiency gains. On ActivityNet-QA~\cite{yu2019activityqa}, reducing tokens from $2056\rightarrow206$ lowers TFLOPs from $11.61\rightarrow1.17$ and reduces latency from $0.447$s to $0.146$s. 


ReDiPrune is plug-and-play across architectures, as it can be inserted between the vision encoder and projector without retraining or model-specific tuning. By shortening the visual sequence early, it reduces latency and accelerates inference across diverse multimodal tasks. Its query-aware selection with diversity regularization also acts as a denoising step, which can lead to small but consistent accuracy gains. Across benchmarks, ReDiPrune is competitive at matched compute, with clearer gains on longer, more redundant videos. For example, with LLaVA-NeXT-Video-7B~\cite{zhang2024llavanextvideo}, retaining only 15\% of visual tokens improves accuracy by 4.6\% on EgoSchema with $2.6\times$ faster inference, and increases WUPS by 0.34\% on NextQA with a $2.7\times$ speedup. Overall, ReDiPrune offers a practical and general solution for scaling real-world multimodal systems without sacrificing task performance.

Our main contributions are summarized as follows:
\begin{itemize}

\item We introduce \textit{ReDiPrune}, a simple, training-free pre-projection token pruning module that preserves fine-grained semantics while improving efficiency.

\item We design a lightweight scoring objective that integrates text-conditioned relevance with max-min diversity, enabling per-frame selection of informative and non-redundant tokens.

\item We show that pre-projection pruning effectively preserves visual semantics and lowers computation. Through ablation studies and experiments on six video and five image benchmarks, ReDiPrune consistently lowers latency while maintaining or improving accuracy, achieving a strong balance between speed and performance.

\end{itemize}

\section{Related Work}
\label{sec:relatedwork}

\subsection{Token Reduction in MLLMs}

Multimodal large language models such as LLaVA~\cite{liu2024llava}, MiniGPT-4~\cite{zhu2023minigpt4}, and Video-ChatGPT~\cite{li2023videochatgpt} couple visual encoders with LLMs, but inference is often bottlenecked by processing dense visual tokens. Existing token reduction methods mainly operate either after projection inside the LLM (post-projection) or before projection in the visual feature space (pre-projection). 

Post-projection approaches typically leverage LLM attention signals: FastV~\cite{chen2024FastV} drops low attention tokens across layers, VTW~\cite{lin2025VTW} halts vision tokens after a calibrated layer via KL guided search, ATP-LLaVA~\cite{ye2024atpllava} learns adaptive layer-wise thresholds, and DART~\cite{wen2025dart} reduces redundancy using pivot based similarity filtering; diversity oriented variants include DivPrune~\cite{Alvar2025divprune} with max min selection and CDPruner~\cite{zhang2025CDPruner} with instruction conditioned diversity. 

Pre-projection methods prune/merge before the projector, including LLaVA-PruMerge~\cite{yang2024llavaprumerge} using ViT attention to merge low salience patches, TRIM~\cite{song2024TRIM} selecting tokens by CLIP~\cite{clip:2021} similarity with adaptive thresholds, PruneVid~\cite{huang2025PruneVid} combining temporal merging with LLM pruning for video, and EfficientLLaVA~\cite{liang2025efficientllava} applying SRM guided search to reduce LLM weights, although projector computation may still remain a bottleneck.

\subsection{Query-Aware and Text-Guided Token Selection}


TRIM~\cite{song2024TRIM} selects query-relevant patches via CLIP~\cite{clip:2021} text-image similarity but only during instruction tuning. ATP-LLaVA~\cite{ye2024atpllava} sets layer-specific pruning thresholds via cross-modal attention. CDPruner~\cite{zhang2025CDPruner} combines instruction relevance and diversity through text-weighted DPP kernels but prunes post-projection. PruneVid~\cite{huang2025PruneVid} ranks tokens by question-to-video attention.  While Q-Adapter~\cite{Chen2025QAdapter} improves query conditioning using adapters rather than reducing tokens. 

In contrast to prompt agnostic strategies such as PruMerge~\cite{yang2024llavaprumerge} and diversity only selection in DivPrune~\cite{Alvar2025divprune}, \textbf{ReDiPrune} performs training-free, query-aware token selection directly in the vision feature space $E_V$ before the projector. By jointly optimizing text relevance and max min diversity at inference time without fine tuning or architectural changes, ReDiPrune enables efficient and context sensitive token reduction while preserving compatibility with existing multimodal LLM pipelines.

\section{Method}
\label{method}

We present \textbf{ReDiPrune}, a training-free, plug-and-play token pruning module for multimodal LLMs. ReDiPrune operates \emph{before} the vision-language projector, selecting a small subset of visual tokens that are both (i) relevant to the user prompt and (ii) diverse within each frame. We first describe where ReDiPrune fits in a standard MLLM pipeline (\S\ref{sec:overview}). Next, we construct a prompt-derived query embedding (\S\ref{sec:query_embedding}), define the relevance/diversity scoring components and the optional candidate pre-filter (\S\ref{sec:scoring}), and finally introduce a greedy solver for the unified objective with the complete procedure in Algorithm~\ref{alg:qprune} (\S\ref{sec:greedy_solver}). We conclude with complexity and practical notes (\S\ref{sec:complexity}).

\subsection{Overview and Problem Setup}
\label{sec:overview}

A multimodal large language model (MLLM) processes video (or image) and text through three stages:
(1) vision encoding, (2) vision-language projection and concatenation, and (3) language decoding.
Given a video with frames $\left\{I^{(n)}\right\}_{n=1}^{N}$ (an image is the special case $N{=}1$), a vision encoder produces per-frame patch tokens
$E_V^{(n)} = \left\{x_i^{(n)}\right\}_{i=1}^{L_n} \in \mathbb{R}^{L_n \times C_v}$.
A multimodal projector $P(\cdot)$ maps them into the LLM embedding space:
$Z_V^{(n)} = P\left( E_V^{(n)} \right) \in \mathbb{R}^{L_n \times H}$.
The projected visual tokens are inserted at the \texttt{<video>} / \texttt{<image>} marker and concatenated with text embeddings $Z_T$ to form the multimodal sequence
$\mathcal{Z}=\left[ Z_T;Z_V \right]$.
A causal language model $f_\theta(\cdot)$ then autoregressively predicts output tokens:
$y_t = f_{\theta}(\mathcal{Z}_{\le t})$, $p\left( y_t|\mathcal{Z}_{\le t} \right)=\mathrm{Softmax}\left( W_o h_t \right)$, where $t$ is the token position, $h_t$ is the hidden state at position $t$ and $W_o$ is the output projection matrix.

\medskip
\noindent\textbf{Challenge:}
Video inputs often produce thousands of visual tokens (frames $\times$ patches per frame), leading to quadratic attention cost, redundant spatio-temporal information, and high memory and latency.

\medskip
\noindent\textbf{Where ReDiPrune is applied:}
ReDiPrune is inserted \emph{between} the vision encoder and the projector. For each frame $n$, it selects a subset of tokens
$\tilde E_V^{(n)} \subseteq E_V^{(n)}$ with 
$\left| \tilde E_V^{(n)} \right| = k_n \ll L_n$, where $L_n$ is the original number of visual tokens produced by the vision encoder for frame $n$ and $k_n$ is the pruning budget. Only $\tilde E_V^{(n)}$ is fed to the projector, yielding $P\left( \tilde E_V^{(n)} \right)$, which reduces computation for both the projector and the LLM while retaining prompt-relevant information.

\subsection{Prompt-Derived Query Embedding}
\label{sec:query_embedding}

\begin{algorithm}[t]
\caption{ReDiPrune}
\label{alg:qprune}
\scriptsize
\SetKwInOut{KwIn}{Input}\SetKwInOut{KwOut}{Output}
\KwIn{Visual features $\{E_V^{(n)}\}$, retention ratio $r$, prompt $p$, $\alpha$, threshold $\tau$, cap $M$, $\beta$}
\KwOut{Pruned features $\{\tilde E_V^{(n)}\}$}

$q \leftarrow \mathrm{BuildQuery}(p)$;\ $\hat q \leftarrow \mathrm{norm}(q)$\;

\For{each frame $n$}{
    $E \leftarrow E_V^{(n)}$;\ $L_n \leftarrow |E|$;\ $R \leftarrow \{1,\dots,L_n\}$;\ $S \leftarrow \emptyset$\;

    $\hat X \leftarrow \mathrm{norm}(E)$;\ $D \leftarrow \mathbf{1}-\hat X\hat X^\top$\;
    $s_i \leftarrow \hat x_i^\top \hat q,\ \forall i\in\{1,\dots,L_n\}$\;
    $k_n \leftarrow \max(1,\mathrm{round}(rL_n))$;\ $M_n \leftarrow \max\!\bigl(k_n,\min(M,L_n)\bigr)$\;
    \lIf{$\tau \le 0$}{$\mathcal{C}\leftarrow R$}
    \Else{
        $\mathcal{C}\leftarrow \{i\in R: s_i\ge \tau\}$\;
        \lIf{$|\mathcal{C}|<k_n$}{$\mathcal{C}\leftarrow \mathrm{top}_{M_n}(R,s)$}
    }
    $\mathcal{C}\leftarrow \mathrm{top}_{\min(|\mathcal{C}|,M_n,\beta k_n)}(\mathcal{C},s)$\;

    $i_0 \leftarrow \arg\max_{i\in \mathcal{C}} s_i$;\ $S \leftarrow S\cup\{i_0\}$;\ $R \leftarrow R\setminus\{i_0\}$\;

    \While{$|S|<k_n$}{
        $K_{\mathrm{rem}}\leftarrow k_n-|S|$;\ $M_{\mathrm{rem}}\leftarrow \max\!\bigl(K_{\mathrm{rem}},\min(M,|R|)\bigr)$\;

        \lIf{$\tau \le 0$}{$\mathcal{C}\leftarrow R$}
        \Else{
            $\mathcal{C}\leftarrow \{i\in R: s_i\ge \tau\}$\;
            \lIf{$|\mathcal{C}|<K_{\mathrm{rem}}$}{$\mathcal{C}\leftarrow \mathrm{top}_{M_{\mathrm{rem}}}(R,s)$}
        }
        $\mathcal{C}\leftarrow \mathrm{top}_{\min(|\mathcal{C}|,M_{\mathrm{rem}},\beta K_{\mathrm{rem}})}(\mathcal{C},s)$\;

        $j^* \leftarrow \arg\max_{i\in \mathcal{C}}\Big[\min_{u\in S} D[u,i]+\alpha s_i\Big]$\;
        $S \leftarrow S\cup\{j^*\}$;\ $R \leftarrow R\setminus\{j^*\}$\;
    }

    $\tilde E_V^{(n)} \leftarrow E[\mathrm{sort}(S)]$\; 
}

\Return{$\{\tilde E_V^{(n)}\}$}
\end{algorithm}

ReDiPrune uses a lightweight, prompt-derived query embedding to guide pruning.
Let the prompt $p$ contain $J$ text tokens. A text encoder produces token embeddings $\left\{ e_j \right\}_{j=1}^{J}$.
We compute a weighted aggregation
\begin{equation}
q \;=\; \sum_{j=1}^{J} w_j\,e_j,
\label{eq:build_query}
\end{equation}
where weights $\{w_j\}$ emphasize informative words ({\em e.g.}, via uniform averaging, self-attention, or position-based weighting).
The resulting vector is aligned to the visual feature dimension $C_v$ (when needed) and $\ell_2$-normalized:
$\hat q \;=\; \mathrm{norm}(q)$.
This yields a query direction in the same feature space as $E_V^{(n)}$, enabling efficient cosine similarity scoring before any cross-modal projection.

\subsection{Scoring Components: Text Relevance and Visual Diversity}
\label{sec:scoring}


Let \(E_V^{(n)}\) denote the vision tokens for frame \(n\), and let \(L_n\) be the number of tokens in that frame. For each frame \(n\), we normalize \(E_V^{(n)}\) as
\begin{equation}
\hat{X}^{(n)} = \mathrm{norm}\!\left(E_V^{(n)}\right)
= \left[\hat{x}_1^{(n)}, \dots, \hat{x}_{L_n}^{(n)}\right].
\end{equation}

\noindent
\textbf{Text relevance:}
Let \(\hat x_i^{(n)}\) denote the \(i\)-th normalized token of frame \(n\). Each token then receives a prompt-conditioned relevance score
\begin{equation}
s_i \;=\; \left\langle \hat x_i^{(n)}, \hat q \right\rangle,\quad i=1,\dots,L_n .
\label{eq:relevance}
\end{equation}
This training-free relevance prior highlights potentially useful regions for answering the query.

\medskip
\noindent
\textbf{Visual diversity:}
To avoid selecting redundant patches, we compute cosine dissimilarity
\begin{equation}
D_{ij} \;=\; 1 - \left\langle \hat x_i^{(n)},\hat x_j^{(n)}\right\rangle,
\qquad D\in\mathbb{R}^{L_n\times L_n}.
\label{eq:diversity}
\end{equation}
Larger $D_{ij}$ indicates tokens capturing more distinct visual content.

\medskip
\noindent
\textbf{Text-conditioned candidate pre-filtering:}
Given threshold $\tau\in[-1,1]$, we define a candidate set
$\mathcal{C}_n(\tau)\;=\;
\bigl\{\, i\in\{1,\dots,L_n\} : s_i\ge\tau \,
\bigr\}.
\label{eq:candidate_set}
$
When $\tau\le 0$, we keep all tokens as candidates.
When $\tau>0$, we restrict candidates to $\mathcal{C}_n(\tau)$; if $|\mathcal{C}_n(\tau)|<k_n$,
we fall back to the top-$M_n$ tokens ranked by $s_i$ to maintain robustness,
where $M_n=\max \, \bigl(k_n,\min \, (M,L_n) \bigr)$ and $M$ is a user-chosen candidate cap. For efficiency, we optionally further limit the candidate pool to top-$\min(M_n,\beta k_n)$ tokens (with $\beta{=}3$).

\subsection{Unified Relevance-Diversity Selection}
\label{sec:greedy_solver}

Given a per-frame budget $k_n=\max\left( 1,\mathrm{round}(rL_n) \right)$, where $r$ is the per frame retention ratio, ReDiPrune selects a subset that is both query-relevant and diverse.
We formalize this as the following objective over index sets $S\subseteq\{1,\dots,L_n\}$:
\begin{equation}
\argmax_{\substack{S\subseteq \{1,\dots,L_n\}\\ |S|=k_n}}
\left(
\min_{\substack{i\neq j\\ i,j\in S}} D_{ij} \;+\; \alpha \sum_{i\in S} s_i
\right),
\label{eq:qprune_obj}
\end{equation}
where $D_{ij}$ and $s_i$ are defined in Eqs.~\eqref{eq:relevance}-\eqref{eq:diversity} and $\alpha$ balances the two terms.
The first term encourages non-redundant coverage (max-min diversity), while the second term promotes alignment with the prompt.
When $\alpha=0$ and $\tau\le 0$, Eq.~\eqref{eq:qprune_obj} reduces to diversity-only max-min selection.

We adopt a greedy solver: initialize the subset with the most relevant token, then iteratively add the token maximizing
\begin{equation}
\mathrm{score}_i \;=\; \min_{u\in S} D_{u i} \;+\; \alpha\, s_i.
\label{eq:greedy_score}
\end{equation}
where $u$ indexes tokens already selected in $S$, and $
i$ indexes remaining candidates. After selection, tokens are restored to their original order (by patch index within each frame and by frame order in video) before projection.

After pruning, only $\tilde E_V^{(n)}$ is passed through the projector $P(\cdot)$ and concatenated with text tokens for decoding. ReDiPrune operates entirely at inference time and requires no model retraining.

\subsection{Complexity and practical integration}
\label{sec:complexity}

ReDiPrune runs independently per frame in the encoder feature space.
For a frame with $L_n$ tokens, computing 
$\hat X^{(n)} \left(\hat X^{(n)} \right)^\top$ costs 
$O \left( L_n^2 C_v \right)$,
and greedy selection costs 
$O \left( k_n \left| \mathcal{C}_n(\tau) \right| \right)$ 
evaluations of Eq.~\eqref{eq:greedy_score}.
In practice, pruning substantially reduces the downstream projector and LLM workload, and sorting the selected indices restores the original spatial-temporal ordering before projection.

\section{Experiment}
\label{exp}

\subsection{Experiment Setting}

\noindent\textbf{Models and baselines:}
We evaluate ReDiPrune using two video backbones, Video-LLaVA-7B~\cite{videollava} and LLaVA-NeXT-Video-7B~\cite{zhang2024llavanextvideo}, and one image backbone, LLaVA-1.5-7B~\cite{liu2024llava}.
For each backbone, we compare against matched baselines: (i) the corresponding unpruned model~\cite{videollava,zhang2024llavanextvideo,liu2024llava}; (ii) DivPrune~\cite{Alvar2025divprune}; and (iii) CDPruner~\cite{zhang2025CDPruner}. For image-only experiments, we also include PruMerge~\cite{yang2024llavaprumerge}, PruMerge$+$~\cite{yang2024llavaprumerge}, TRIM~\cite{song2024TRIM}, and DART~\cite{wen2025dart}.

We adopt DivPrune and CDPruner as our primary training-free baselines. DivPrune performs diversity-based token selection, while CDPruner uses text-guided scoring. Other training-free pruning methods, such as FastV~\cite{chen2024FastV} and VTW~\cite{lin2025VTW}, have been extensively evaluated in the DivPrune and CDPruner studies and do not outperform them under comparable settings; we therefore focus on these stronger recent alternatives.
For image evaluation, we also include DART~\cite{wen2025dart}, LLaVA-PruMerge~\cite{yang2024llavaprumerge}, and TRIM~\cite{song2024TRIM} as baselines. We note that TRIM, DART, and LLaVA-PruMerge do not release code for video models; accordingly, our main study targets video settings, where token redundancy is substantial and thus provides a realistic testbed for pre-projection pruning.

\medskip
\noindent\textbf{Benchmarks:} 
We evaluate 4 video-language and 5 image-language understanding tasks.
The video suite includes NextQA~\cite{xiao2021next}, EgoSchema~\cite{Mangalam2023EgoSchema}, ActivityNet-QA~\cite{yu2019activityqa}, and Video-ChatGPT~\cite{li2023videochatgpt}.
The image suite includes GQA~\cite{hudson2019GQA}, MMB~\cite{liu2024mmbench}, MME~\cite{fu2024mme}, POPE~\cite{li2023POPE}, and ScienceQA-IMG (SQA)~\cite{lu2022SQA}.

\medskip
\noindent\textbf{Metrics and protocols:} 
We report accuracy and LLM-assisted scores where applicable. For video-language tasks, ActivityNet-QA is evaluated with both accuracy and an LLM judge using standardized prompts; EgoSchema uses accuracy only; Video-ChatGPT uses the LLM judge only; and NextQA is evaluated with Wu-Palmer similarity (WUPS)~\cite{Wu1994WUPS}.
For image-language tasks, GQA and SQA report Exact Match (EM), MMB reports Accuracy, MME reports Perception Score (P-score), and POPE reports F1.
When required, LLM-based metrics are computed via the ChatGPT API using the official instruction templates of each benchmark to ensure consistency across models and pruning methods. More details about these metrics are explained in Supplementary.

To demonstrate the superiority of computational efficiency, we report the computational requirements, measured in TFLOPs, for the baseline and comparable pruning methods. For a decoder layer with hidden dimension $d$ and FFN intermediate dimension $m$, the dominant floating-point operations (FLOPs) at sequence length $n$ are approximated by $g(n) = 4nd^{2} + 2n^{2}d + 2ndm$. For a decoder with $T$ layers, the total cost without pruning is $ \mathrm{Full} = T\,g(n_{\mathrm{full}})$. If token pruning takes effect after the first $K$ layers, the cost becomes $\mathrm{Pruned}(K) = K\,g(n_{\mathrm{full}}) + (T-K)\,g(n_{\mathrm{pruned}})$. We report the TFLOPs ratio $
\frac{\mathrm{Pruned}(K)}{\mathrm{Full}},
$ where lower values indicate higher efficiency. The sequence lengths before and after pruning are
$
n_{\mathrm{full}} = t + v_{\mathrm{full}}$ and
$n_{\mathrm{pruned}} = t + v_{\mathrm{pruned}}
$,
where $t$ denotes the number of text tokens and $v_{\mathrm{full}}$ / $v_{\mathrm{pruned}}$ denote the numbers of visual tokens before and after pruning. Here $T$ is the number of decoder layers, and $K$ is the index of the first layer operating on the shorter sequence ($K{=}0$ applies pruning before the decoder, while $K{=}T$ recovers the baseline). We report TFLOPs as $\mathrm{FLOPs}/10^{12}$.


\medskip
\noindent\textbf{Implementation details:}
All experiments are conducted on 4$\times$A100\;80\,GB GPUs with a batch size of 1.
We run benchmark evaluations using \texttt{lmmsevals}~\cite{zhang2024lmms} and compute LLM-based metrics through real-time API calls to \texttt{gpt-3.5-turbo}.
On short-form datasets (e.g., NextQA), the pre-filter threshold shows negligible sensitivity; therefore we set $\tau{=}0$ by default and focus on the relevance-diversity trade-off controlled by $\alpha$ and $r$ (see Supplementary for details).

\subsection{Video-Language Understanding}

In this section, we evaluate our proposed ReDiPrune on two LLaVA-based video-language models, Video-LLaVA-7B~\cite{videollava} and LLaVA-NeXT-Video-7B~\cite{llavavideo}, to assess performance across diverse video-language tasks (Table~\ref{tab:Results_Video-LLaVA-7B}). Experiments are mainly conducted on four datasets: ActivityNet-QA~\cite{yu2019activityqa}, NextQA~\cite{xiao2021next}, EgoSchema~\cite{Mangalam2023EgoSchema}, and Video-ChatGPT~\cite{li2023videochatgpt}. 
DivPrune~\cite{Alvar2025divprune} and CDPruner~\cite{zhang2025CDPruner} are selected as comparable baselines, representing the most recent advances in visual token pruning. All results are reproduced under our unified experimental setting for fair comparison. Additionally, more video-language understanding experiments can be found in Supplementary. 

\begin{table*}[t]
\caption{Results on video-language benchmarks with Video-LLaVA-7B~\cite{videollava} and LLaVA-NeXT-Video-7B~\cite{zhang2024llavanextvideo}. Left: NextQA/EgoSchema at keep ratio 0.15 (TFLOPs column 1). Right: ActivityNet-QA/Video-ChatGPT at keep ratio 0.10 (TFLOPs column 2). ReDiPrune achieves the best accuracy-efficiency trade-off.}
\label{tab:Results_Video-LLaVA-7B}
\centerline{
\resizebox{1.0\linewidth}{!}{
\setlength{\tabcolsep}{1.2mm}
\renewcommand{\arraystretch}{1.5}
\begin{tabular}{c|c|c|c|c||c|c|c}
    \hline
     & \textbf{Method} &
    \makecell{\textbf{NextQA}~\cite{xiao2021next}\\WUPS~$\uparrow$} & \makecell{\textbf{EgoSchema}~\cite{Mangalam2023EgoSchema}\\Acc~$\uparrow$} &
    \makecell{\textbf{TFLOPs}\\(ratio \%)} &
    \makecell{\textbf{ActivityNet-QA}~\cite{yu2019activityqa}\\Acc / Score~$\uparrow$} & \makecell{\textbf{Video-ChatGPT}~\cite{li2023videochatgpt}\\Score~$\uparrow$} & 
    \makecell{\textbf{TFLOPs}\\(ratio \%)} \\
    \hline
    \modelgroup{4}{LLaVA 1.5-7B}
    & \cellcolor{rowgray}Original*       & \cellcolor{rowgray}15.22 & \cellcolor{rowgray}38.2 & \cellcolor{rowgray}12.177 (100)    & \cellcolor{rowgray}42.73 / 3.33 & \cellcolor{rowgray}2.13 & \cellcolor{rowgray}11.606 (100) \\
    & DivPrune*~\cite{Alvar2025divprune} & 15.60                    & 41.4                    & 2.188 (17.86)                      & 44.63 / 3.37                    & 2.18                    & 1.174 (10.11) \\
    & CDPruner*~\cite{zhang2025CDPruner} & 15.40                    & 42.4                    & 2.188 (17.86)                      & 44.86 / 3.37                    & 2.12                    & 1.174 (10.11) \\
    & \textbf{ReDiPrune }            & \textbf{15.92}           & \textbf{43.2}           & 2.188 (17.86)                      & \textbf{45.69 / 3.40}           & \textbf{2.24}           & 1.174 (10.11) \\
    \hline
    \modelgroup{4}{\makecell{LLaVA-NeXT\\-Video-7B}}
    & \cellcolor{rowgray}Original*       & \cellcolor{rowgray}26.33 & \cellcolor{rowgray}43.6 & \cellcolor{rowgray}29.918 (100)    & \cellcolor{rowgray}44.76 / 3.04 & \cellcolor{rowgray}2.520 & \cellcolor{rowgray}29.211 (100) \\
    & DivPrune*~\cite{Alvar2025divprune} & 23.70                    & 38.8                    & 4.429 (14.77)                      & 42.09 / 2.97                    & 2.637                   & 2.712 (9.28) \\
    & CDPruner*~\cite{zhang2025CDPruner} & 23.53                    & 39.2                    & 4.429 (14.77)                      & 41.19 / 2.91                    & 2.629                   & 2.712 (9.28) \\
    & \textbf{ReDiPrune }            & \textbf{26.42}           & \textbf{45.6}           & 4.429 (14.77)                      & \textbf{44.83 / 3.06}           & \textbf{2.663}          & 2.712 (9.28) \\
    \hline
\end{tabular}%
}
}
\end{table*}

For the Video-LLaVA-7B backbone, we compare the original model, DivPrune, CDPruner, and our ReDiPrune. All pruning methods substantially reduce computational cost from 11.89 TFLOPs to 1.41 TFLOPs (11.8\% of the original) while maintaining or improving performance. DivPrune relies on diversity-based token selection, and CDPruner introduces text-conditioned scoring. In contrast, ReDiPrune jointly models visual diversity and text relevance within a unified per-frame pruning framework, delivering the strongest overall performance. It achieves the highest WUPS on NextQA (\textbf{15.92}), the best accuracy on EgoSchema (\textbf{43.2}), the top ActivityNet-QA results (\textbf{45.69 / 3.40}), and the highest Video-ChatGPT score (\textbf{2.24}). These results demonstrate that ReDiPrune removes redundant visual tokens without sacrificing, and in fact improving, multimodal reasoning quality, yielding a superior accuracy-efficiency trade-off under identical computational budgets.

For the LLaVA-NeXT-Video-7B backbone, ReDiPrune again have the most balance between performance and efficiency. While reducing computation from 29.92 TFLOPs to 4.43 TFLOPs (14.77\% of the original), it consistently achieves the best results across all four benchmarks. Specifically, ReDiPrune improves NextQA WUPS over the unpruned baseline (\textbf{26.42} vs.\ 26.33) and substantially increases the precision of EgoSchema (\textbf{45.6} vs.\ 43.6). It also achieves the highest Video-ChatGPT score (\textbf{2.663}) and the strongest ActivityNet-QA performance (\textbf{44.83 / 3.06}), outperforming both DivPrune and CDPruner by clear margins. Overall, these results show that ReDiPrune strikes a strong balance between efficiency and multimodal understanding. By jointly optimizing visual diversity and text relevance at the frame level before projection, it removes redundant tokens while preserving semantic fidelity, supporting stronger reasoning across video-language benchmarks.

\begin{figure*}[t]
\centerline{
\includegraphics[width=1.0\textwidth]{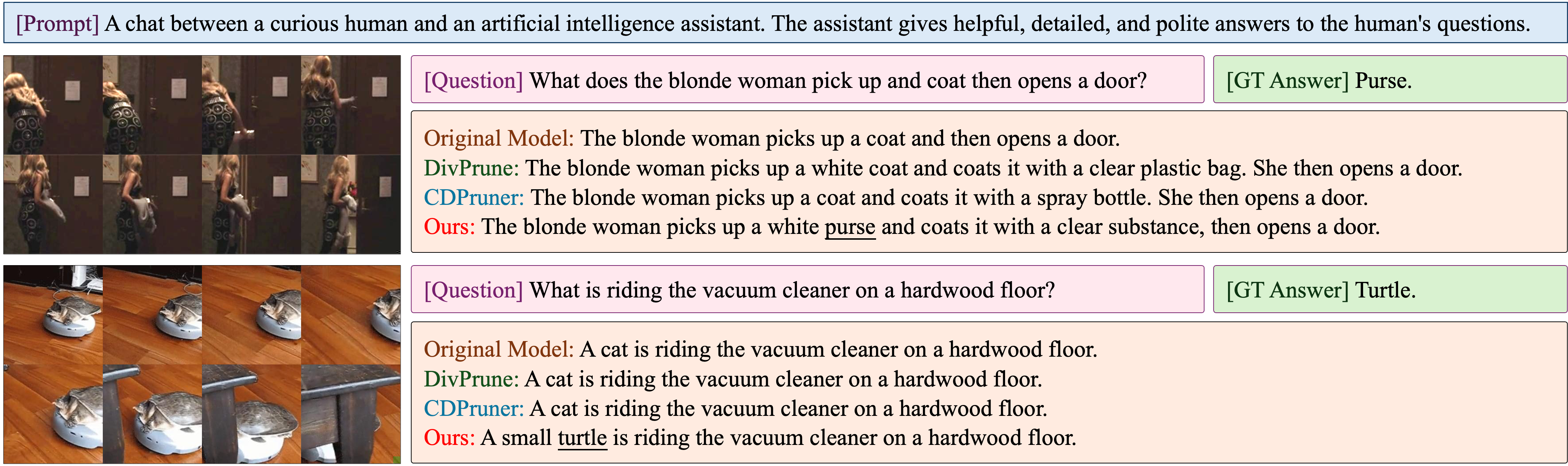}
}  
\caption{\textbf{Qualitative examples from the TGIF dataset~\cite{Jang2017TGIF} using Video-LLaVA-7B~\cite{videollava}.} For each question, we show the ground-truth (GT) answer and responses from the original model, DivPrune~\cite{Alvar2025divprune}, CDPruner~\cite{zhang2025CDPruner}, and ReDiPrune. ReDiPrune accurately captures action cues, demonstrating stronger semantic grounding and temporal understanding compared with competing pruning methods.}
\label{fig:qualitative_video_tgif}
\vspace{-2mm}
\end{figure*}

\begin{table}[t]
\caption{Results on image-language benchmarks using the LLaVA-1.5-7B~\cite{liu2024llava} backbone show that ReDiPrune delivers strong accuracy even at a strict keep ratio of $r = 0.1$, demonstrating the effectiveness of its pre-projection pruning strategy. Note that PruMerge$+$~\cite{yang2024llavaprumerge} operates with a larger effective token budget. TRIM**~\cite{song2024TRIM} is implemented in a training-free manner in our own setup, while TRIM*~\cite{song2024TRIM} is reproduced with $r=0.1$ following the instruction-tuning procedure described in the original paper. }
\label{tab:Results_LLaVA-1.5-7B_image}
\vspace{-2mm}
\centerline{
\resizebox{1.0\linewidth}{!}{
    \begin{tabular}{c|c|c|c|c|c||c|c}
        \hline
        \textbf{Method} &  
        \makecell{\textbf{GQA}~\cite{hudson2019GQA}\\EM~$\uparrow$} & 
        \makecell{\textbf{MMB}~\cite{liu2024mmbench}\\Acc~$\uparrow$} & 
        \makecell{\textbf{MME}~\cite{fu2024mme}\\P-score~$\uparrow$} & 
        \makecell{\textbf{POPE}~\cite{li2023POPE}\\F1~$\uparrow$} &
        \makecell{\textbf{SQA}~\cite{lu2022SQA}\\EM~$\uparrow$} &
        \makecell{\textbf{Visual} \\ \textbf{Token} $\#$} &
        \makecell{\textbf{TFLOPs}\\(ratio \%)} \\ \hline
        \rowcolor{rowgray}
        Original* & 61.9 & 64.7 & 1506.5 & 85.9 & 69.5 & 576 & 2.23 (100) \\
        PruMerge*~\cite{yang2024llavaprumerge} & 51.79 & 54.89 & 1218.13 & 64.2 & 67.82 & 58 & 0.53 (16.13) \\
        PruMerge$+$*~\cite{yang2024llavaprumerge} & 59.45 & 63.23 & 1439.12 & 82.88 & 67.72 & 230 & 1.41 (43.53) \\
        TRIM*~\cite{song2024TRIM} & 39.07 & 58.16 & 1074.31 & 71.15 & 67.65 & 58 & 0.53 (16.13) \\
        TRIM**~\cite{song2024TRIM} & 54.17 & 59.44 & 1261.33 & 83.99 & 68.17 & 58 & 0.53 (16.13) \\
        DART*~\cite{wen2025dart} & 55.77 & 60.90 & 1378.54 & 71.85 & 68.96 & 58 & 0.53 (16.13) \\
        DivPrune*~\cite{Alvar2025divprune} & 57.36 & 59.45 & 1361.62 & 85.39 & 68.32 & 58 & 0.54 (16.42) \\
        CDPruner*~\cite{zhang2025CDPruner} & 57.21 & 58.42 & 1355.85 & 82.14 & 68.52 & 58 & 0.53 (16.13)  \\ \hline
        \textbf{ReDiPrune} & \textbf{57.62} & \textbf{59.88} & \textbf{1392.90} & \textbf{85.39} & \textbf{69.11} & 58 & 0.53 (16.29)    \\ \hline
    \end{tabular}
    }
}
\vspace{-1mm}
\end{table}

\medskip
\noindent\textbf{Qualitative Results:}
We present qualitative comparisons on TGIF dataset~\cite{Jang2017TGIF} using the Video-LLaVA-7B~\cite{videollava} model, as shown in Fig.~\ref{fig:qualitative_video_tgif}. Each example displays the input prompt, question, ground-truth answer, and model responses from the original unpruned model and other pruned methods.


In the first example, the unpruned and baseline-pruned models focus on the coat, whereas ReDiPrune correctly identifies the \textit{purse}, which appears in only two frames, and produces a coherent answer that matches the ground truth. This improvement indicates that ReDiPrune better preserves query-relevant visual evidence.
In the second, all non-ReDiPrune models predict \textit{cat}, likely due to coarse shape/color cues. Conversely, ReDiPrune accurately recognizes the entity as a \textit{turtle}, showcasing its capacity to preserve nuanced yet semantically significant visual indicators.
Overall, these qualitative findings demonstrate that ReDiPrune facilitates enhanced visual grounding and semantic accuracy, producing responses that are both more precise and more closely aligned with the intended meaning of the inquiry.

\subsection{Image-Language Understanding}

We evaluate ReDiPrune on image-language understanding using LLaVA-1.5-7B~\cite{liu2024llava} across five benchmarks: GQA~\cite{hudson2019GQA}, MMBench~\cite{liu2024mmbench}, MME~\cite{fu2024mme}, POPE~\cite{li2023POPE}, and ScienceQA-IMG~\cite{lu2022SQA}.
Comparisons are made against the unpruned LLaVA-1.5-7B and six pruning baselines: PruMerge~\cite{yang2024llavaprumerge}, PruMerge$+$~\cite{yang2024llavaprumerge}, TRIM~\cite{song2024TRIM}, DART~\cite{wen2025dart}, DivPrune~\cite{Alvar2025divprune}, and CDPruner~\cite{zhang2025CDPruner}. 
As shown in Table~\ref{tab:Results_LLaVA-1.5-7B_image}, ReDiPrune achieves strong performance across all benchmarks while strictly maintaining same token budget. Some baselines, such as PruMerge$+$, operate with a larger token budget and are therefore not directly comparable under this constraint. 

On the POPE dataset, ReDiPrune attains an F1 score of 85.39, matching DivPrune and remaining within 0.5 of the unpruned model (85.9). On the SQA dataset, it achieves the highest EM score (69.11) outperforming all pruning methods ({\em e.g.}, 68.32 for DivPrune, 68.52 for CDPruner). ReDiPrune also achieves the best GQA EM (57.62), the highest MME P-score (1392.90), and competitive accuracy on MMBench (59.88). 
Although DART reports a slightly higher MMBench score (60.90), it uses a different automatic judge version (our \texttt{gpt-3.5-turbo-0125} vs. \texttt{gpt-3.5-turbo-0613}), which may explain the difference.

While PruMerge$+$ performs competitively ({\em e.g.}, 82.88 on POPE, 1439.12 on MME), it does not strictly adhere to the same pruning ratio. Its results rely on retaining all spatially sampled and aggregated tokens in addition to the main selections, raising the effective token budget to roughly 18-30\% depending on input image resolution. In contrast, the original PruMerge follows the strict $r=0.1$ setting and shows correspondingly lower performance (e.g., 64.2 on POPE, 1218.13 on MME).

For TRIM~\cite{song2024TRIM}, we report two settings to separate the effect of instruction tuning from pruning. TRIM* follows the original instruction-tuning procedure with TRIM at $r=0.1$, while TRIM** is our training-free implementation at the same keep ratio. TRIM** consistently improves over TRIM* and provides a stronger training-free baseline, but ReDiPrune still achieves higher accuracy on every benchmark in Table~\ref{tab:Results_LLaVA-1.5-7B_image} under the same strict token budget. In particular, ReDiPrune have the best accuracy on all five image datasets, while matching TRIM’s compute cost. Overall, ReDiPrune achieves the strongest accuracy-efficiency trade-off under strict token budgets, demonstrating robust performance and generalizability across diverse image-language tasks.
\begin{table}[t]
\caption{Efficiency comparison. ActivityNet-QA~\cite{yu2019activityqa} results for Video-LLaVA-7B~\cite{videollava} and LLaVA-NeXT-Video-7B~\cite{zhang2024llavanextvideo}, and five datasets average for LLaVA-1.5-7B~\cite{liu2024llava}. Metrics: prefill time, end-to-end (E2E) latency, and max GPU memory.}
\label{tab:Ablation}
\vspace{-2mm}
\centerline{
\resizebox{1.0\linewidth}{!}{
    \begin{tabular}{c|c|c|c||c|c|c||c|c|c}
        \hline
        & \multicolumn{3}{c||}{\textbf{Video-LLaVA-7B}} & \multicolumn{3}{c||}{\textbf{LLaVA-NeXT-Video-7B}} & \multicolumn{3}{c}{\textbf{LLaVA-1.5-7B}} \\ \cline{2-4}\cline{5-7}\cline{8-10}
        \multirow{2}{*}{\textbf{Method}} & 
        \makecell{\textbf{Prefill}\\\textbf{Time} (sec)} & \makecell{\textbf{E2E Latency}\\(Sec) $\downarrow$} & \makecell{\textbf{Max GPU}\\mem (GB) $\downarrow$} &
        \makecell{\textbf{Prefill}\\\textbf{Time} (sec)} & \makecell{\textbf{E2E Latency}\\(Sec) $\downarrow$} & \makecell{\textbf{Max GPU}\\mem (GB) $\downarrow$} &
        \makecell{\textbf{Prefill}\\\textbf{Time} (sec)} & \makecell{\textbf{E2E Latency}\\(Sec) $\downarrow$} & \makecell{\textbf{Max GPU}\\mem (GB) $\downarrow$} \\ \hline
        \rowcolor{rowgray}
         Original*                          & 0.369 & 0.447 & 29.92    & 0.395 & 0.526 & 16.54   & 0.069 & 0.201 & 13.23 \\
         DivPrune*~\cite{Alvar2025divprune} & 0.089 & 0.124 & 27.96    & 0.126 & 0.163 & 13.76   & 0.048 & 0.137 & 13.23 \\
         CDPruner*~\cite{zhang2025CDPruner} & 0.134 & 0.166 & 27.96    & 0.215 & 0.250 & 13.76   & 0.055 & 0.140 & 13.23 \\ \hline
         \textbf{ReDiPrune}            & 0.114 & 0.146 & 27.96    & 0.169 & 0.205 & 13.76   & 0.053 & 0.137 & 13.23 \\ \hline
    \end{tabular}
    }
}
\vspace{-2mm}
\end{table}
\begin{table}[t]
\caption{Ablation study of different query weighting strategies in ReDiPrune on the ActivityNet-QA~\cite{yu2019activityqa} benchmark using LLaVA-NeXT-Video-7B~\cite{llavavideo}.
}
\label{tab:ablation1}
\centerline{
\begin{tabular}{c|c|c}
  \hline
  \textbf{Method} & \textbf{Accuracy} & \textbf{Score} \\ \hline
  Prune w/o text                    & 44.4625 & 3.0566 \\
  Prune w/ average weighting        & 44.6000 & 3.0624 \\
  Prune w/ middle-peak weighting    & 44.7375 & 3.0640 \\
  Prune w/ self-attention weighting & 44.7500 & 3.0651 \\
  Prune w/ exponential weighting    & \textbf{44.8250} & 3.0636 \\ \hline
\end{tabular}
}
\vspace{-2mm}
\end{table}


\begin{figure*}[t]
\centerline{
    \includegraphics[width=\linewidth]{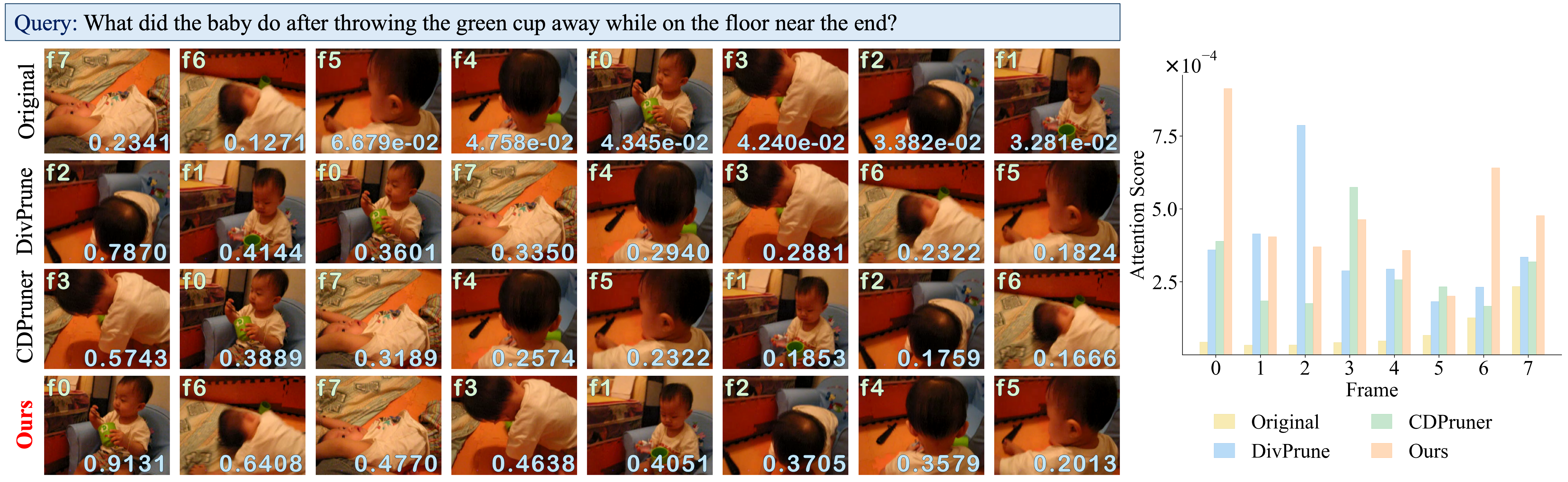}        
}
\caption{\textbf{Sharper, query-aligned attention with ReDiPrune.} Frame-wise attention visualization and distribution for different pruning methods on a NextQA~\cite{xiao2021next} sample. \textbf{(Left)} Each method (Original, DivPrune~\cite{Alvar2025divprune}, CDPruner~\cite{zhang2025CDPruner}, and ReDiPrune) processes the same query; eight frames (f0-f7) and their attention scores (in $\times 10^{-3}$) are shown in descending order of importance. \textbf{(Right)} Histogram of attention across frames. ReDiPrune produces sharper, query-focused attention on the most relevant frames and a more concentrated distribution, indicating stronger alignment with query.}
\label{fig:attention} 
\vspace{-1mm}
\end{figure*}


\subsection{Efficiency Analysis}

Table~\ref{tab:Ablation} reports computational efficiency on ActivityNet-QA~\cite{yu2019activityqa} for the two video backbones (Video-LLaVA-7B~\cite{videollava} and LLaVA-NeXT-Video-7B~\cite{zhang2024llavanextvideo}), and the average efficiency over five image datasets for the LLaVA-1.5-7B~\cite{liu2024llava} image backbone. We compare prefill time, end-to-end (E2E) latency, and maximum GPU memory usage.

On ActivityNet-QA, pruning consistently reduces prefill time and end-to-end (E2E) latency relative to the original model while also lowering peak GPU memory for both video backbones. For Video-LLaVA-7B, E2E latency drops from 0.447s (Original) to 0.124s (DivPrune), 0.166s (CDPruner), and 0.146s (ReDiPrune); for LLaVA-NeXT-Video-7B, it decreases from 0.526s to 0.163s, 0.250s, and 0.205s, respectively. Peak memory reduces from 29.92GB to 27.96GB on Video-LLaVA-7B and from 16.54GB to 13.76GB on LLaVA-NeXT-Video-7B, reflecting the lighter visual-token workload after pruning.
DivPrune attains the lowest prefill time and E2E latency because it is text-agnostic and avoids text-conditioned scoring overhead. In contrast, CDPruner and ReDiPrune incorporate text-guided scoring, which introduces a small runtime cost but better preserves semantically relevant visual evidence. Despite this modest overhead, ReDiPrune remains efficient (0.146s on Video-LLaVA-7B and 0.205s on LLaVA-NeXT-Video-7B) while delivering stronger task performance, as reflected by the higher benchmark scores in Table~\ref{tab:Results_Video-LLaVA-7B}.

For LLaVA-1.5-7B, the reported numbers are averaged over 5 image datasets, where single-image inputs exhibit lower token redundancy than videos. Accordingly, pruning yields smaller but consistent latency improvements (0.201s to 0.137s) and modest prefill reductions (0.069s to 0.053s), while peak memory remains unchanged at 13.23GB across methods.

Table~\ref{tab:Ablation} shows that pre-projection pruning primarily accelerates inference by reducing prefill and total decoding time, with the largest gains in video models where token redundancy is substantial. ReDiPrune achieves efficiency comparable to DivPrune/CDPruner while delivering stronger accuracy on video benchmarks (Table~\ref{tab:Results_Video-LLaVA-7B}), supporting a trade-off between accuracy and efficiency.

\subsection{Ablation Study and Analysis}

\textbf{Effect of Pruning:} As shown in Table~\ref{tab:Results_Video-LLaVA-7B}, ReDiPrune consistently improves accuracy and WUPS across all benchmarks comparing to other pruning methods. In contrast, on the LLaVA-NeXT-Video-7B backbone~\cite{zhang2024llavanextvideo}, both DivPrune and CDPruner underperform unpruned model on NextQA~\cite{xiao2021next} and EgoSchema~\cite{Mangalam2023EgoSchema}.
This highlights that applying either diversit-only or relevance-conditioned after projection pruning may remove useful contextual cues, whereas our unified approach better preserves the task-relevant visual semantics.
To probe why pruning helps, we examine attention over video frames on a NextQA sample (see Fig.~\ref{fig:attention} for the query). Fig.~\ref{fig:attention} visualizes the attention maps and their frame-wise histograms. In the original model, attention is broadly distributed across redundant frames, including those where the baby remains still or background objects dominate, thereby diluting the model’s ability to focus on the action of interest.
DivPrune narrows this spread but lacks semantic guidance, sometimes emphasizing visually distinct yet irrelevant frames;
CDPruner adds text conditioning but tends to overfit to locally relevant regions, neglecting global temporal cues.
In contrast, ReDiPrune allocates stronger attention to the key action frames ($f0$-$f3$), where the baby picks up and interacts with the cup, while suppressing redundant frames later in the sequence.
This selective focus aligns visual attention with the linguistic query, leading to more coherent reasoning and accurate predictions.

Overall, pruning in ReDiPrune acts as an implicit regularizer: 
by removing redundant or noisy visual tokens, it enhances the model’s ability to align and ground visual evidence to the query, 
thereby improving multimodal reasoning performance even under reduced token budgets.

\medskip
\noindent\textbf{Effect of Query Weighting:} Table~\ref{tab:ablation1} examines different query-weighting strategies for generating the query embedding $q$ on ActivityNet-QA using LLaVA-NeXT-Video-7B. Removing text conditioning leads to a clear drop in performance, showing that visual pruning must be guided by the question. Uniform and average weighting offer small gains by adding basic semantic signals, whereas middle-peak and self-attention schemes perform better by highlighting tokens near the main action words. The exponential weighting strategy achieves the best accuracy (\textbf{44.83\%}) and score (\textbf{3.0636}), suggesting that gradually increasing weights toward the end of the query helps capture key verbs and entities that appear late in instruction-style prompts. Overall, the results indicate that exponential weighting better highlights informative words, leading to sharper text-guided pruning and improved multimodal reasoning.

\section{Conclusion}
\label{conclusion}

We introduced \textbf{ReDiPrune}, a training-free pre-projection token pruning method for multimodal large language models. By jointly modeling text-conditioned relevance and visual diversity directly in the vision encoder feature space, ReDiPrune selects a compact and representative subset of visual tokens before cross-modal projection. This design preserves fine-grained semantics while substantially reducing downstream computation. Across multiple video and image benchmarks, ReDiPrune consistently achieves a strong accuracy-efficiency trade-off under strict token budgets, often improving task performance while significantly reducing TFLOPs and latency. 

\medskip
\noindent
\textbf{Limitation \& Future Work:} Our experiments focus on LLaVA-style architectures and frame-wise selection. Extending ReDiPrune to more diverse multimodal backbones and to globally optimized spatio-temporal pruning remains future work. In addition, adaptive token budgeting strategies may further improve robustness across tasks and input complexity.

{
    \small
    \bibliographystyle{splncs04}
    \bibliography{main}
}


\end{document}